\documentclass[a4paper]{jpconf}
\usepackage{graphicx}

\begin{document}
\title{Obstacle Detection Quality as a Problem-Oriented Approach to Stereo Vision Algorithms Estimation \\ in Road Situation Analysis}

\author{A.A. Smagina, D.A. Shepelev, E.I. Ershov, A.S. Grigoryev}

\address{Institute for Information Transmission Problems (Kharkevich Institute) --IITP RAS, \\Bolshoy Karetny per. 19, build.1, Moscow, Russia, 127051}

\ead{smagina@visillect.com}

\begin{abstract}
In this work we present a method for performance evaluation of stereo vision based obstacle detection techniques that takes into account the specifics of road situation analysis to minimize the effort required to prepare a test dataset. This approach has been designed to be implemented in systems such as self-driving cars or driver assistance and can also be used as problem-oriented quality criterion for evaluation of stereo vision algorithms.
\end{abstract}

\section{Introduction}

One of the purposes of any traffic situation analysis system, including driver assistance and fully operational autopilot, is avoidance of collisions with both static and moving objects. 
Currently the most common means for solving this task are radars, lidars and various stereo vision systems (including binocular stereo cameras, time-of-flight cameras, and structured light scanners).  \cite{Mukhtar, Franke, Shepelev}
Due to relatively low price and high spatiotemporal resolution binocular cameras are especially interesting as being limited primarily by the difficulty of 3D reconstruction which is a rapidly advancing field of study. They also are passive sensors, which is an advantage in under crowded conditions. Many new binocular stereo vision algorithms are published each year (e.g. \cite{Pang}, \cite{Saleem}) and it is important to evaluate quality of both collision avoidance algorithm as a whole and the underlying stereo disparity reconstruction in particular.
Having sufficiently precise 3D world model the task of detecting and avoiding obstacles can be straightforwardly formalized thus allowing exact, rather than fuzzy heuristic verification. 

Binocular reconstruction algorithms are usually tested by the Middlebury dataset and toolset.
 \cite{Scharstein}. However, since it is impossible to create a general stereo matching algorithm for all practical purposes, problem-oriented criteria are gaining more attention \cite{Sidorchuk}. 
In particular, paper \cite{Geiger} describes the popular KITTI dataset for testing road situation analysis algorithms and demonstrates that performance ranking using KITTI is drastically different compared to Middlebury.
Hence, for developing of obstacle detection systems one needs datasets and metrics specifically designed for road situation analysis. For instance, in the traffic sign recognition problem there are datasets with corresponding problem-oriented evaluation methods already proposed \cite{shakhuro}.

However, using correct datasets is not sufficient for adequate applicability evaluation of 3D reconstruction algorithms. One also needs to analyze the impact of 3D reconstruction on the final outcomes, i.e. to perform a higher-level analysis.
The authors of \cite{Schneider} suggest a three-level hierarchy of quality metrics for stereo vision algorithm evaluation:

\begin{itemize}
\item Low level -- correctness of point cloud and depth map -- evaluates the precision of the reconstruction algorithm itself but gives little information about the quality of the final problem and demands time-consuming preparation of highly detailed reference 3D data.

\item Medium level -- the so called ``stixel'' representation -- the 3D world is presented by a simplified geometrical model allowing to work with reconstruction results on a level with less details yet relevant enough for the task being solved, thus simplifying the preparation of reference data.

\item High level -- the final answer of the system under test. In \cite{Schneider} this was the position of the nearest vehicle, however the metric itself was mentioned rather superficially.
\end{itemize}

Apart from inability to produce the overall system quality, middle level metrics also have a costly procedure of obtaining test data (ground truth). 

The paper \cite{Schneider} studies the method of obtaining the correct answers in the form of stixel representation from radars or lidars followed by automatic conversion from depth maps or point clouds into stixel representation. However such conversion demands manual stixel editing to correct the ground truth still yielding high labor cost of the method.

High labor cost is not an issue when using cameras with typical position relative to the vehicle and when target observation conditions are present in the existing open datasets, for example KITTI \cite{Geiger} and Cityscapes \cite{Marius}.
However when using more exotic configurations like the one described in this paper (self-driving bus --- cameras are mounted high and tilted downwards), one needs to prepare a new dataset. In such cases there is an issue of minimizing the cost of its preparation while maintaining satisfactory detalization relatively to the problem being solved by the system.

This paper considers obstacles quality evaluation as an estimation of the final answer the obstacle avoidance system generates, e.g. was decision to stop a vehicle generated correctly or not. Ground truth for such final answers are generated from low-detailed obstacles descriptions and do not lead to high cost while collecting a dataset. This scheme corresponds to medium and high abstraction levels in the \cite{Schneider} hierarchy and can be used for testing both the system as a whole (with a fixed decision taking module) and as a problem oriented comparison metric for stereo vision algorithms. The method can easily be generalized for systems which use other types of sensors providing an optical and depth image. Quality estimation of obstacle tracking is beyond the scope op this work.

To introduce the problem context we provide an example of the obstacle detection system which was the initial target of the suggested quality evaluation approach.
We study the approach to quality evaluation itself and present an example of using the described method for both optimizing the parameters of 3D reconstruction algorithm and analyzing the system behavior as a whole.

\section{Obstacle detection system}

\begin{figure}[h]
    \begin{center}
        \includegraphics[width=1.0\linewidth]{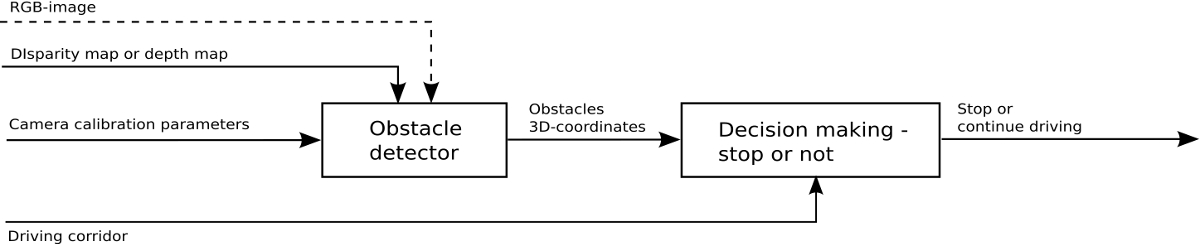}
    \end{center}
    \caption{\label{ODScheme}Obstacle detection system architecture.}
\end{figure}

Let us describe the architecture of the obstacle detection system (see fig. \ref{ODScheme}).
It uses the disparity or depth map, RGB image (optionally), sensor calibration parameters and the information about driving corridor provided by the trajectory planning subsystem as its input. It is assumed that the radial distortion of input RGB images is already compensated.  
The correction might be performed with the use of calibration images or via blind radial distorsion compensation  \cite{kunina}.
This modular structure allows several options for the depth sensor: visible light stereo systems, time-of-flight cameras, structured light systems, etc.

In this work we obtained images from binocular camera with known calibration parameters mounted on a vehicle looking in direction of the motion positioned about 2.2 meters above the ground level.
The depth map is computed using the stereo matching algorithm with subsequent filtering of the obtained disparity map. Implementation details are given below in section 4.

The system output is a \textit{binary answer indicating obstacle presence or absence}.
Thus the suggested scheme can be used both as a framework for prototyping an obstacle detection system and as a high level abstraction of a more sophisticated system that models the decision making procedure.

An important feature of such system is the possibility to consider the stereo matching algorithm errors with different weights depending on object distance.
Indeed, in our task the insensitivity to distant objects has no importance, while a failure to detect a closely located object may be fatal and the proximity measure is defined by the target properties of the autonomous system in development.

\section{Quality Evaluation}

\begin{figure}[!h]
    \begin{center}
        \includegraphics[width=1.0\linewidth]{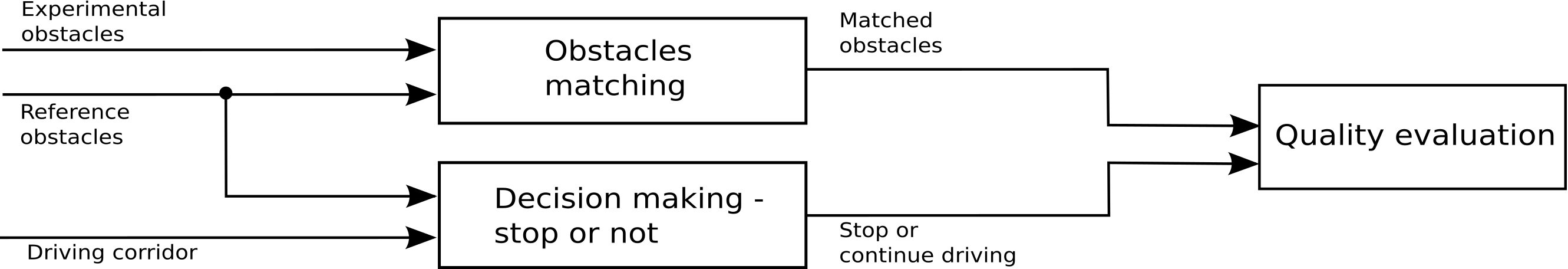}
    \end{center}
    \caption{\label{QEScheme}Structural scheme of obstacle detection quality evaluation with respect to stops.}
\end{figure}

Testing of the proposed obstacle detection system implies \textit{estimation of vehicle stops correctness}  (see fig. \ref{QEScheme}).

Ground truth annotations should be provided as 3D-obstacles coordinates (x, y marked as bounding rectangle, z estimated from calibration) to avoid boundary effects. 
Any three-dimensional object can be represented by a bounding box or a set of stixels \cite{Pfeiffer}. 
However, generation of such markup is labor-intensive and can be simplified without loss of quality for the target task as will be shown later.
We suggest representing the image markup and the result of obstacle detection by rectangles in a vertical plane. It is sufficient then to prepare the scene markup only in a problem-specific visibility zone, since activation of obstacle detector are relevant primarily for the vehicle driving corridor.

In our work the object markup (further referred as \textit{marked obstacle}) is performed for single images from the stereo camera, obtained by virtual camera transformation to a viewpoint perpendicular to the road surface by means of a rectangle with vertical and horizontal sides.
 
\begin{figure}[!h]
    \begin{minipage}[h]{0.49\linewidth}
        \center{\includegraphics[width=0.95\linewidth]{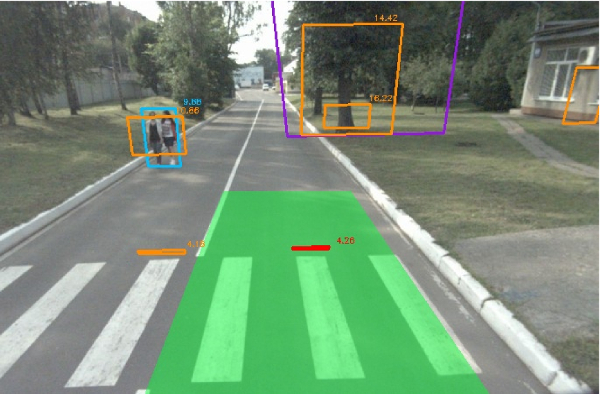} \\ a)}
    \end{minipage}
    \hfill
    \begin{minipage}[h]{0.49\linewidth}
        \center{\includegraphics[width=0.95\linewidth]{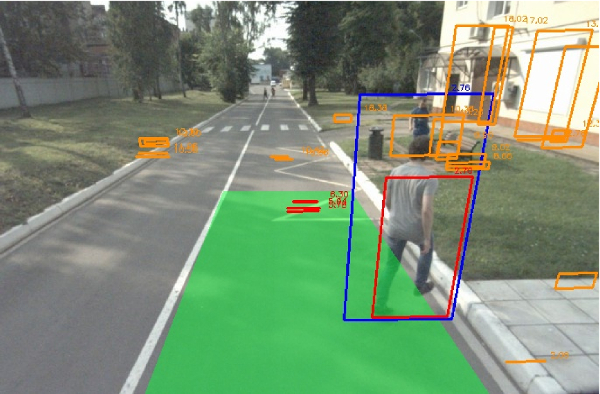} \\ b)}
    \end{minipage}
    \caption{\label{Matching} The results of detected obstacles matching (marked red and orange inside and outside the driving corridor resp.) with the marked ones (dark and light blue inside and outside the driving corridor resp.): false-positive stop (a), true-positive stop (b). The driving corridor is shown green, zones of indifference are violet.}
\end{figure}

All large objects-obstacles including people, vehicles, walls, fences are marked in this manner.
In fig. \ref{Matching}a and \ref{Matching}b the marked obstacles are represented by rectangles.
Additional \textit{zones of indifference} can be marked (violet polygons in fig.\ref{Matching}a) for the case when we do not expect a confident obstacle detection, in presence of a mesh fence or bushes.
Zones of indifference are usually needed in working with a wide driving corridor.
Knowing intrinsic and extrinsic camera parameters one can compute 3D coordinates of the source 3D space box of the marked rectangle thus computing distance to it.

A similar representation can be obtained for the obstacle detector answer: for all outstanding objects a planar quadrangle is constructed which is tangent to the object front part and parallel to the image plane (further referred as \textit{detected obstacles}).

In the proposed scheme shown in fig. \ref{QEScheme} the obstacles marked on a frame are proposed to be matched against the detected obstacles. The matched answers are counted only relatively to the obstacles falling inside the driving corridor.

Each marked obstacle is considered as matched with detected obstacles if it satisfied two conditions. The first condition restricts distances to detected and marked obstacles by the following relation:

$$\frac{|z_{ref} - z_{exp}|}{z_{ref}} < \mathrm{T},$$  

where $z_{ref}$ is the distance to the marked obstacle, $z_{exp}$ is the distance to the detected obstacle, \mbox{$\mathrm{T}$ is the threshold} value (typically 0.25).
The second condition is that the intersection of marked and detection obstacles rectangles should be non-empty.

If detected obstacle cannot be matched with any marked one, including zones of indifference, then it is considered as a false positive obstacle. 
If marked obstacle cannot be matched with any of detected obstacle, it considered to be false negative, otherwise it is considered as a true positive obstacle.

The correctness of vehicle stop is estimated by the following rule:

\begin{itemize}
\item the stop is considered true positive if the driving corridor contains at least one true positive obstacle (see fig. \ref{Matching}b),
\item the stop is considered false positive, if the driving corridor contains only false positive obstacle (see fig. \ref{Matching}a),
 \item the stop is considered false negative, if the driving corridor contains only false negative obstacle,
\item otherwise the stop is considered true positive.
\end{itemize}

\section{Experiment setup}

Data acquisition was performed using a binocular camera mounted on a vehicle at 2.2 m height tilted $20^{\circ}$ downwards relatively to the horizon.
The camera view angle was \mbox{about $80^{\circ}$}.
The output image size was $1280 \times 1024$ pixels. The stereo base width was 75 cm.

\begin{figure}[h]
    \begin{center}
        \includegraphics[width=1.0\linewidth]{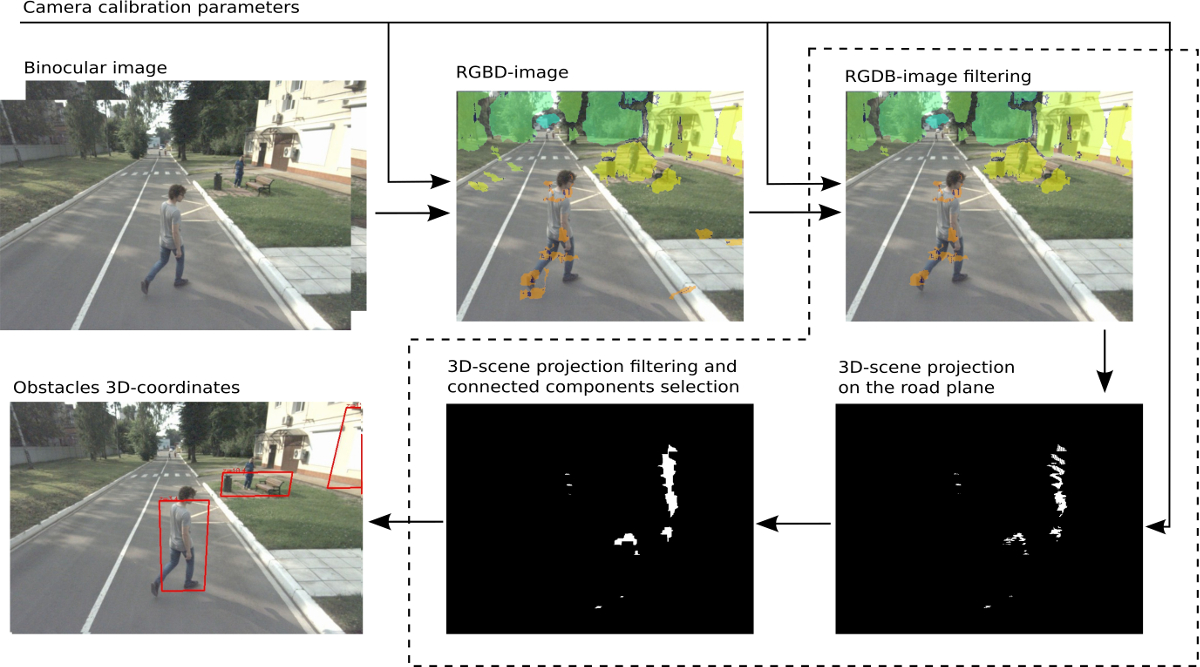}
    \end{center}
    \caption{\label{OD} Stage based demonstration of obstacle detector operation. Dashed blocks are directly related to the obstacle detector.}
\end{figure}

To demonstrate the applicability of the proposed quality evaluation method we implemented an obstacle detection algorithm which is schematically shown in fig. \ref{OD}.

The disparity map was constructed using the StereoBM local stereo matching algorithm from the OpenCV library \cite{Faugeras} with subsequent conversion to the the depth map. The depth map is fed as input to the obstacle detector filtered by spatial coordinates to cut out the road plane (presuming that the road plane cannot be tilted by more than $10^{\circ}$ relatively to the current vehicle position).

The road plane cutoff height is an adjustable parameter of the obstacle detector.
The 3D scene projection on the road plane is then constructed as a single channel image which is also further filtered (here we used morphological closure) and then on the obtained binary image connected components are extracted and the corresponding 3D coordinates are retrieved.
The connected components obtained in this manner can be filtered by their area.

The filter parameters are adjustable and demand optimization on the marked dataset to increase the target overall quality metric.

\section{Evaluation results}

The sequences we used for evaluation were recorded with the 3 Hz frequency and were consisted of 5-20 frames.
In total the dataset contains 1794 frames, 65\% of which were recorded in simple conditions (day light, no precipitation) and 35\% in difficult conditions (rainy weather, night lighting, rooms with light reflecting floor), 840 frames contain marked obstacles in the driving corridor. We checked the detections of the obstacle detection system in the driving corridor of 2.5 m width and 7 m length.

\begin{figure}[!h]
    \includegraphics[width=0.5\linewidth]{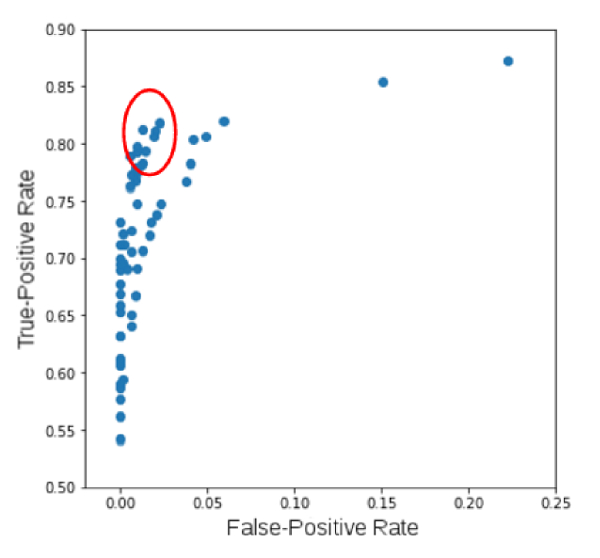}\hspace{2pc}
    \begin{minipage}[b]{14pc}\caption{\label{Optim} Example of parameters optimization process. Every single point on a TPR-FPR plot corresponds to a single point in parameter space. With red ellipse the optimal parameters according to project technical task are marked.}\end{minipage}
\end{figure}

To test the suggested quality assessment method we optimized the system by the stereo algorithm and/or the obstacle detection algorithm parameters. 
So we produced a grid in a parameter space and for each point of this grid we calculated true positive stops rate (TPR) and false positive stops rate (FPR). 
An example of TPR vs FPR plot obtained in a such way is shown at the fig. \ref{Optim}. 
From that plot we choose optimal values of parameters that satisfied projects technical goal.  

\begin{figure}[!]
    \begin{minipage}[h]{0.49\linewidth}
        \flushright{\includegraphics[width=0.65\linewidth]{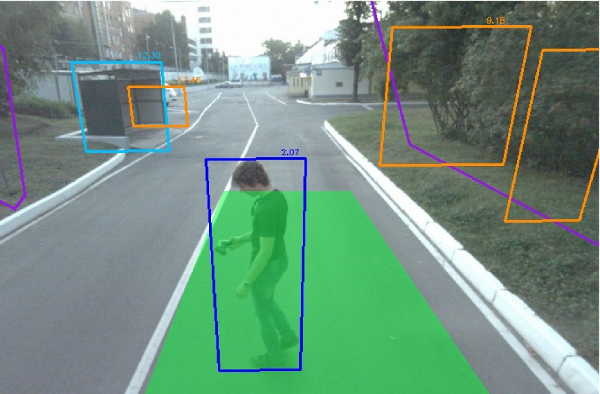}}
    \end{minipage}
    \hfill
    \begin{minipage}[h]{0.49\linewidth}
        \flushleft{\includegraphics[width=0.65\linewidth]{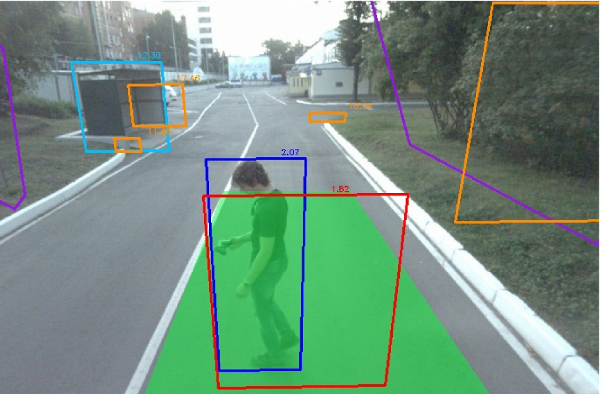}}
    \end{minipage}
    \vfill
    \begin{minipage}[h]{0.49\linewidth}
        \flushright{\includegraphics[width=0.65\linewidth]{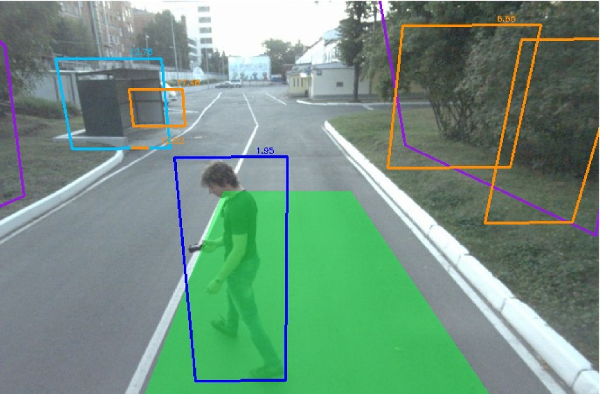}}
    \end{minipage}
    \hfill
    \begin{minipage}[h]{0.49\linewidth}
        \flushleft{\includegraphics[width=0.65\linewidth]{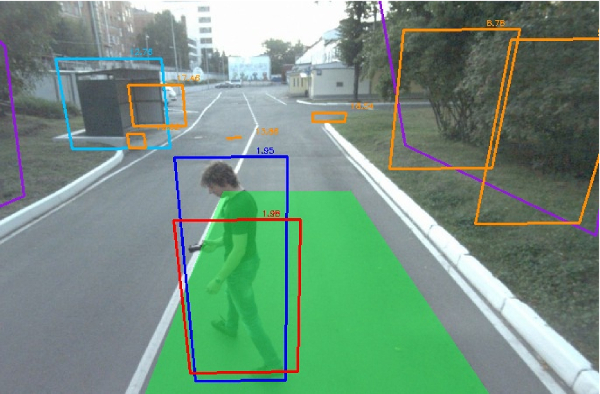}}
    \end{minipage}
    \vfill
    \begin{minipage}[h]{0.49\linewidth}
        \flushright{\includegraphics[width=0.65\linewidth]{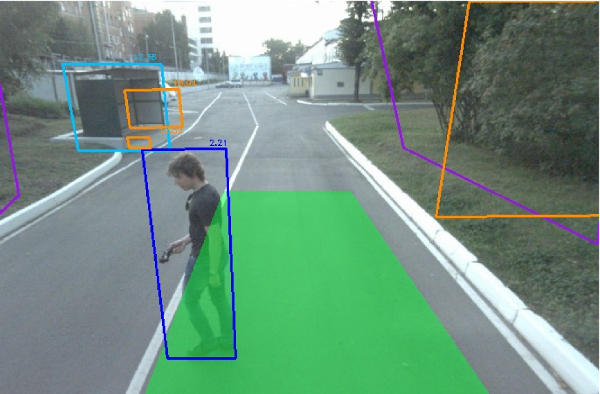}}
    \end{minipage}
    \hfill
    \begin{minipage}[h]{0.49\linewidth}
        \flushleft{\includegraphics[width=0.65\linewidth]{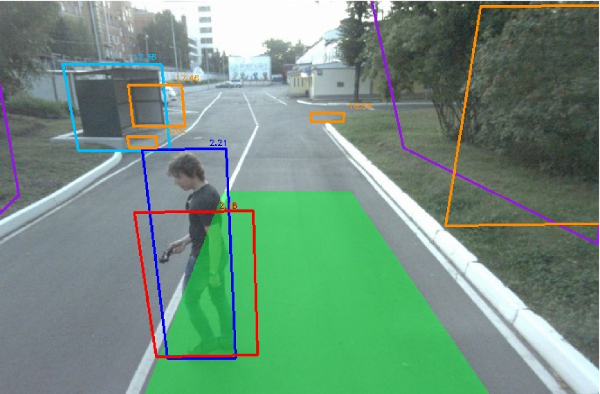}}
    \end{minipage}
    \center{a)}\vfill
    \vspace{0.2cm}
    \vfill
    \begin{minipage}[h]{0.49\linewidth}
        \flushright{\includegraphics[width=0.65\linewidth]{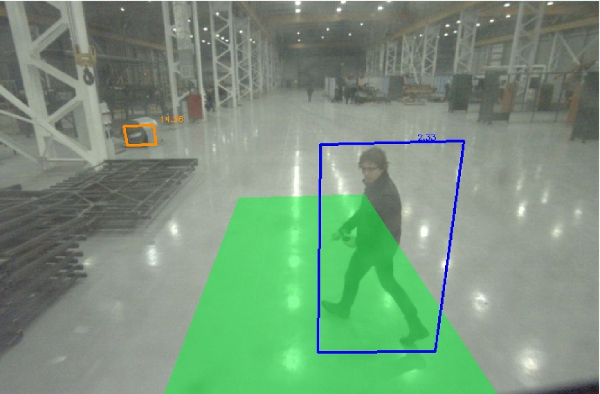}}
    \end{minipage}
    \hfill
    \begin{minipage}[h]{0.49\linewidth}
        \flushleft{\includegraphics[width=0.65\linewidth]{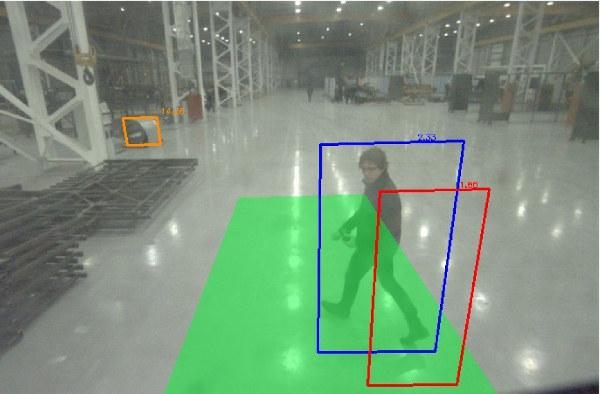}}
    \end{minipage}
    \vfill
    \begin{minipage}[h]{0.49\linewidth}
        \flushright{\includegraphics[width=0.65\linewidth]{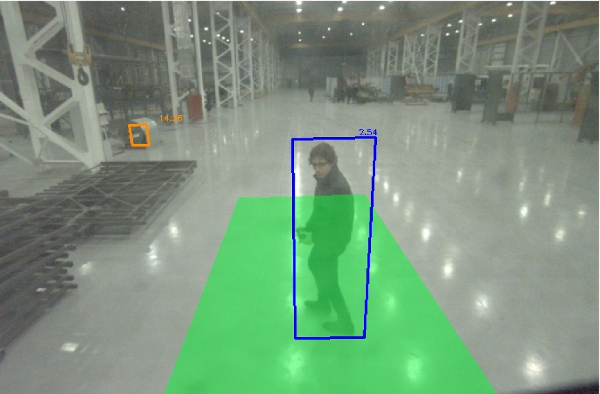}}
    \end{minipage}
    \hfill
    \begin{minipage}[h]{0.49\linewidth}
        \flushleft{\includegraphics[width=0.65\linewidth]{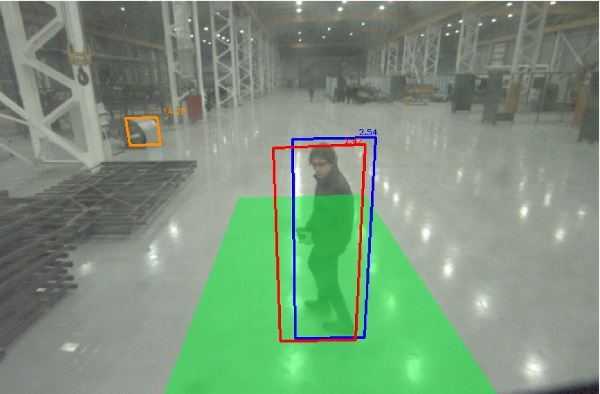}}
    \end{minipage}
    \vfill
    \begin{minipage}[h]{0.49\linewidth}
        \flushright{\includegraphics[width=0.65\linewidth]{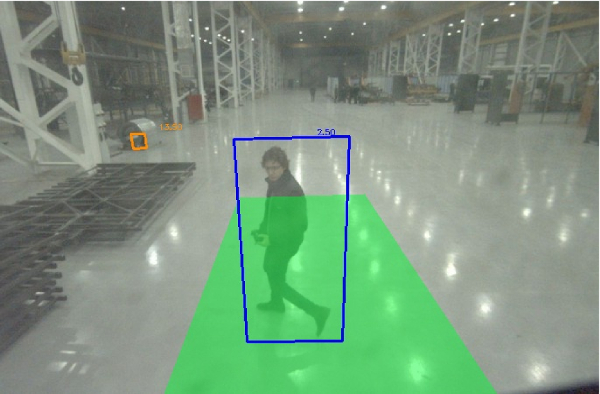}}
    \end{minipage}
    \hfill
    \begin{minipage}[h]{0.49\linewidth}
        \flushleft{\includegraphics[width=0.65\linewidth]{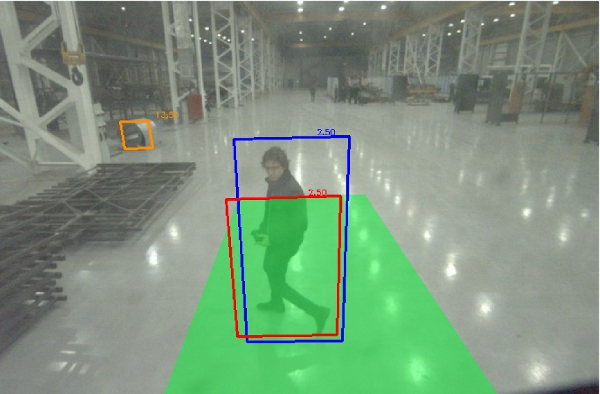}}
    \end{minipage}
    \center{b)}
    
    \caption{\label{Res} Visualization of obstacle detection results in simple (a) and difficult (b) conditions in the example of a three-frame sequence. The results of detection with expert-adjusted parameters are on the left, the results after optimizing the obstacle detector and stereo algorithm parameters are on the right. }
\end{figure}

The results of the optimization process that described above are presented in table \ref{Table} and fig. \ref{Res}. 
As shown in table \ref{Table}, in this task the obstacle detector and stereo algorithm parameters optimization can produce a higher overall system quality compared to adjusting only the stereo matching algorithm. 
The optimization of obstacle detector parameters also produces comparable quality to stereo algorithm optimization but it is easier to compute since it has fewer parameters.

\begin{table}[!b]
    \caption{\label{Table} Obstacle detection quality with different parameter values.}
    \begin{center}
        \begin{tabular}{lll}
        \br
        &True-positive & False-Positive
        \\& stops ratio & stops ratio\\
        \mr
        Initial settings provided by an expert  & 0.677 & 0.000 \\
        Optimization of obstacle detector       & 0.801 & 0.003 \\
        Optimization of stereo algorithm        & 0.819 & 0.013 \\
        Optimization of both                    & 0.822 & 0.011 \\
        \br
        \end{tabular}
    \end{center}
\end{table}

As shown in fig. \ref{Res}, proposed method for quality evaluation allows to tune parameters to be able to drive in a given driving corridor even in complex conditions with multiple reflections from the ground.

\section{Conclusion}

In this paper we study development of a stereo vision based obstacle detectors suitable for traffic situation analysis.
The primary novelty of the described approach is the use of a quality metrics based on a high-level metric working with the final system outcome, i.e. required stops.
We presumed this outcome to be binary but the proposed scheme can be expanded for the case of multiple answers, for example, when the obstacle detection subsystem is provided with decisions about avoiding an obstacle.
The proposed metric can serve both as a robustness characteristic for the obstacle detector as a whole and as a problem-oriented quality evaluation for stereo vision algorithms in various problems of road situation analysis.
The ground truth preparation in the dataset was reduced to object bounding boxes and indifference zone marking, thus having low labor cost and not demanding specific hardware.

To demonstrate the proposed method applicability we have shown a specific implementation of the obstacle detection algorithm which even though simple demonstrates pretty good applicability even in complex conditions.
Moreover, the suggested system can employ the obstacle detection algorithm based on stixel representation allowing to take into account road defects and to monitor the change in obstacle positions.

\setcounter{secnumdepth}{0} 
\section{Acknowledgments}

This research was supported by Russian Science Foundation grant \mbox{(No.~14-50-00150)}.

\setcounter{secnumdepth}{1} 

\end{document}